\pdfoutput=1

\documentclass[11pt]{article}

\usepackage[]{ACL2023}

\usepackage{times}
\usepackage{booktabs}
\usepackage{latexsym}
\usepackage{longtable}
\usepackage{multirow}
\usepackage{rotating}
\usepackage{tikz}

\usetikzlibrary{shapes,arrows}
\usepackage[T1]{fontenc}
\usepackage{tikz}
\usetikzlibrary{arrows.meta}
\usepackage[utf8]{inputenc}
\usepackage{comment}
\usepackage{colortbl}
\usepackage{microtype}
\usepackage{pifont}
\usepackage{inconsolata}

\usepackage{graphicx}

\title{Milestones in Bengali Sentiment Analysis leveraging Transformer-models: Fundamentals, Challenges and Future Directions}

\author{Saptarshi Sengupta, Shreya Ghosh, Prasenjit Mitra \\ The Pennsylvania State University, USA \\ \texttt{\{sks6765,shreya,pmitra\}@psu.edu}
\AND
Tarikul Islam Tamiti \\ Rajshahi University of Engineering \& Technology, Bangladesh \\ \texttt{tarik2568@gmail.com}}

\begin{document}
\maketitle
\begin{abstract}
Sentiment Analysis (SA) refers to the task of associating a \textit{view polarity} (usually, positive, negative, or neutral; or even fine-grained such as slightly angry, sad, etc.) to a given text, essentially breaking it down to a supervised (since we have the view labels apriori) classification task. Although heavily studied in resource-rich languages such as English thus pushing the SOTA by leaps and bounds, owing to the arrival of the Transformer architecture, the same cannot be said for resource-poor languages such as Bengali (BN). For a language spoken by roughly 300 million people, the technology enabling them to run trials on their favored tongue is severely lacking. In this paper, we analyze the SOTA for SA in Bengali, particularly, Transformer-based models. We discuss available datasets, their drawbacks, the nuances associated with Bengali i.e. what makes this a challenging language to apply SA on, and finally provide insights for future direction to mitigate the limitations in the field.

\end{abstract}

\section{Introduction}
Sentiment Analysis, an interdisciplinary field combining computational linguistics, text mining, and machine learning, has grown exponentially over the past decade. It has emerged as a powerful tool for understanding human emotions and opinions in numerous contexts, from product reviews to social media chatter~\cite{yadollahi2017current}. However, much of this growth has focused on resource-rich languages, such as English. There remains, thus, a significant gap in research pertaining to low-resource languages, which are often overlooked due to the lack of readily available linguistic tools and datasets.

Among these low-resource languages, Bengali, is a language of considerable interest. Spoken by over 300 million people globally, it remains underrepresented in SA/NLP in general~\cite{shammi2023comprehensive}. This underrepresentation is problematic as the rich cultural and socio-linguistic context of Bengali users is largely uncaptured, thus limiting our understanding and modeling of their sentiments.

SA in low-resource languages, particularly in Bengali, presents multiple challenges due to a variety of factors such as,

\textbf{Lack of Annotated Corpora} The most significant challenge is the scarcity of large, high-quality, and diverse annotated datasets in Bengali. These corpora are essential for training and validating SA models~\cite{sen2022bangla}. Unlike languages such as English, where multiple large-scale annotated datasets exist, Bengali lacks such resources.

\textbf{Linguistic Complexity} Bengali is a morphologically rich language with complex verb forms, compound words, and a high degree of inflection. This richness adds an extra layer of complexity when developing SA models. Additionally, the language's orthographic variation and lack of standardized spelling further complicate text-processing tasks.

\textbf{Sarcasm and Idiomatic Expressions} Bengali, like many other languages, is rich in idiomatic expressions and sarcasm. Detecting sentiment in such cases requires a deep understanding of the language and cultural context, which is challenging for automated systems.

\textbf{Limited NLP Tools} There is a lack of comprehensive Natural Language Processing (NLP) tools for Bengali. Tools for tokenization, POS-tagging, stemming, and NER, which are readily available for languages like English, are still underdeveloped in Bengali.

Addressing these challenges necessitates a multi-faceted approach; creation of large, diverse, and annotated datasets, the development of more sophisticated NLP tools for Bengali, and the application of culturally sensitive modeling techniques.

This paper presents a comprehensive survey of the current state of Bengali SA, aiming to highlight the challenges and opportunities associated with this low-resource language. It investigates the existing methods and resources in the field and discusses the challenges faced by researchers, such as the lack of un/labelled corpora, linguistic peculiarities of the language, and the cultural nuances that affect sentiment expression. Furthermore, this paper outlines potential strategies and future directions for improving SA in Bengali, thereby fostering a more inclusive representation of global languages in SA research. Through this survey, we aim to provide a solid foundation for researchers venturing into the field of Bengali SA, sparking new ideas, and encouraging the development of more robust, culturally sensitive SA models.

\subsection{Types of SA}

SA, also known as opinion mining, involves using text analysis, natural language processing, and computational linguistics to identify and extract subjective information from source materials. This process is employed in a variety of contexts, and the results can serve various purposes. Although different problems in their own right, we classify SA into the following categories to provide as much as coverage for Bengali as possible,

\emph{Hate Speech Detection:} This SA variant is essential for identifying and flagging language that is deemed offensive, aggressive, or harmful~\cite{romim2021hate}. Its primary application is in moderating online platforms to promote healthy digital communication. Algorithms used in hate speech detection discern between harmless opinion expressions and damaging or toxic language.

\emph{Stance Detection:} This form of SA is intended to establish the stance or viewpoint of a speaker or writer on a specific subject~\cite{roy2021unsupervised}. The stance could be supportive, opposed, or neutral. Stance detection aids in assessing public sentiment concerning controversial subjects, brand opinion, or political inclination.

\emph{Emotion Mining:} This SA category moves beyond simple positive or negative sentiment detection. It aims to pinpoint specific emotions such as joy, anger, sadness, fear, surprise, and so forth, as expressed in text. Emotion mining gives a more profound insight into user reactions to products, services, or events~\cite{iqbal2022bemoc}.

\emph{Aspect-Based SA (ABSA):} ABSA offers a refined version of SA that doesn't just label an entire document as positive, negative, or neutral~\cite{ahmed2021bangla}. Instead, it dissects the sentiment conveyed about different aspects or attributes within the text. For example, in a product review, ABSA can differentiate the sentiment towards the product's cost from the sentiment towards its features or usability.

\emph{Depressive Texts:} This is a specialized branch of sentiment analysis~\cite{hasib2023depression,ghosh2023attention}, that identifies and extracts subjective information from source text. This specialized task focuses on identifying signs of depression, based on the sentiments expressed in a user's written text. 

\subsection{Issues prevalent in SA}

Each type of SA serves different needs and has its unique challenges. For instance, (a) \textbf{Distinguishing hate speech} from other forms of communication can be challenging. Satire, sarcasm, and cultural nuances can often be misinterpreted by algorithms. Furthermore, the definition of what constitutes hate speech can vary significantly across different cultures and legal systems, making it even more challenging to build universally applicable models. (b) One of the main challenges in stance detection is dealing with \textbf{implicit stances}, where the opinion isn't directly stated. Additionally, subjectivity and bias in labeling stance data can affect the accuracy of the model. Also, detecting the stance in less structured texts, such as social media posts, can be challenging due to the use of slang, emojis, and non-standard grammar. (c) Emotion mining is complicated by the fact that the \textbf{same text can evoke different emotions in different people}. Also, people express their emotions in varied ways, making the training data diverse and complex. The use of sarcasm, irony, and other forms of figurative language can lead to misinterpretation of the expressed emotion. (d) \textbf{Challenges in ABSA include accurately identifying aspects or features mentioned in the text}, especially when they are implicitly stated or referred to using different terminology. Also, determining sentiment toward a specific aspect can be complex when the overall sentiment of the text is different or when multiple sentiments are expressed in the same sentence. (e) \textbf{Handling the cultural nuances}, idioms, and language-specific expressions can be tricky. Translation can be used as a workaround, but this often leads to a loss of sentiment-bearing nuances.

Addressing these challenges requires continuous advancements in natural language processing, a deeper understanding of human emotions and language nuances, and the development of more sophisticated machine learning models. The choice of which to use depends on the specific goals of the analysis, the nature of the data at hand, and the resources available for the task.

\subsection{Related surveys}

While there exist several surveys for SA in Bengali viz. \citet{shammi2023comprehensive, sen2022bangla, hira2022systematic, banik2019survey, alam2021review}, none of them truly provide broad coverage of Transformers for multilingual or Bengali applications. Our paper attempts to remedy this by discussing all (to the best of our knowledge) transformer variants applicable to Bengali text and showcasing their performance for the same. For an insight into pre-Transformer era techniques, we refer readers to any one of the above articles.

\subsection*{Datasets and Benchmarks} 

Table~\ref{tab:dataset} provides a comprehensive summary of various research studies benchmarking distinct Natural Language Processing (NLP) tasks related to sentiment analysis. These tasks span a range of sub-domains, including aspect-based sentiment analysis, emotion classification, hate speech detection, and depressive text detection. Each task has been examined within various contexts, such as food reviews, newspaper comments, e-commerce comments, and more.

The sentiment analysis category consists of a total number of samples ranging from approximately 1,000 to over  1,58,065. Model performance varies, with accuracy and F1 scores ranging from around 66\% to above 94\%. In the subdomain of aspect-based sentiment analysis, the tasks have been performed within contexts such as cricket and restaurant reviews. The total sample sizes for these studies sit at around 2,800 to 2,900. The F1 scores for these tasks are reported at 37\% and 42\%, respectively. Emotion classification tasks encompass a variety of contexts, including comments on government policies, socio-political issues, and YouTube comments. The reported accuracy and F1 scores for these studies are approximately 65\% and 62\% respectively.
Hate speech detection tasks deal with detecting hate speech and cyberbullying, with F1 scores reported at 87\% and 85\%, respectively. Lastly, depressive text detection involves categorizing social texts into depressive and non-depressive classes, and models in this domain have achieved an accuracy of 94\%.

The Transformer-based model, BanglaBERT, exhibited superior performance across various domains, notably outperforming other models by achieving a remarkable Weighted Average F1-score of 0.9331~\cite{kabir2023banglabook}. Furthermore, BanglaBERT achieved state-of-the-art performance on the SentNoB dataset~\cite{bhattacharjee-etal-2022-banglabert}. In the context of emotion classification, the Bangla-BERT model achieved a macro average F1-score of 24.61 across 22,698 Bangla public comments from social media platforms, covering 12 different domains~\cite{islam2022emonoba}, which is a quite difficult dataset. Lastly, Transformer-based models, BERT and ELECTRA, were deployed effectively for hate-speech detection, achieving  accuracies of 85.00\% and 84.92\%, respectively, highlighting their potential in large-scale sentiment analysis tasks~\cite{aurpa2022abusive}.


\begin{table*}[htbp]
  \centering
  \caption{Dataset Summary and Benchmark. Cl.:   number of classes in the dataset, 
T.S.:total number of samples,
A: Accuracy score in percentage,
F1:F1-score in percentage,
DT: Depressive text detection}
\label{tab:dataset}
  \resizebox{1.0\textwidth}{!}{
  \begin{tabular}{@{}cp{3cm}p{8cm}ccp{2cm}@{}}
    \toprule
     \textbf{Task} & \textbf{Paper} & \textbf{Context} & \textbf{Cl.} & 
     \textbf{T.S.} 
     & \textbf{Result}  \\
    \midrule
     {\multirow{20}{*}{\rotatebox[origin=c]{90}{\textbf{Sentiment Analysis}}}} & \citet{junaid2022bangla} & \emph{(Food Review)} Customer Reviews about the quality of food from online food services (e.g., Foodpanda). Manually labeled as positive and negative  & 2 & 
     1040  
     & A: 90.86 \\
     & \citet{sharmin2021attention} & \emph{(Social Media)} Reviews and comments on books, people, hotels, products, research, events etc.  Manually labeled into three categories (positive, negative, neutral)  & 3 & 2979 
     & A:66.06; F1:66.02 \\
     & \citet{kabir2023banglabook} & \emph{(BanglaBook)} Bangla book reviews classified into three broad categories: positive, negative, and neutral & 3 & 158065 & F1:93.31 \\ 
     & \citet{haydar2018sentiment} & \emph{(Facebook Comments)} Collected from e-commerce and restaurant FaceBook posts and categorized into positive, negative, and neutral class  & 3 & 34271 
     & A: 80 \\
     & \citet{islam2021sentnob} & \emph{(SentNoB)} Comments on 13 different topics from Prothom Alo (Bangla newspaper) articles, YouTube videos, and social media & 3 & 15728 
     &  F1:72.89~\cite{bhattacharjee-etal-2022-banglabert}\\
    & \citet{jabin2022comparison} & \emph{(e-commerce Website)} Reviews collected from Rokomari.com. Labelled into positive and negative class  &  2 & 6652 
    & A:94.5; F1:94.52 \\
& \citet{al2017sentiment} & \emph{(Microblogging Website)} Comments collected. Labelled as positive, negative classes by 500 annotators. 
 & 2 & 16000 
& A:75.5 \\
\midrule
{\multirow{8}{*}{\rotatebox[origin=c]{90}{\textbf{Aspect Based SA}}}}  & \citet{rahman2018datasets} & \emph{(Cricket)} Comments from online sources on five aspect categories (batting, bowling, team, team management, other). Labelled into three sentiment polarities (positive, negative, neutral) and 5 aspect categories & 3, 5 & 2900 
& F1:37 \\
& \citet{rahman2018datasets} & 
\emph{(Restaurant)} Manually translated English restaurant dataset  \cite{pontiki2016semeval} into Bangla. Labelled into five aspect categories (food, price, service, ambiance, miscellaneous) and three sentiment polarities  & 3, 5 & 2800 
& F1:42 \\ 
\midrule
{\multirow{10}{*}{\rotatebox[origin=c]{90}{\textbf{Emotion Classification}}}} & \citet{rahman2019comparison} & \emph{(Facebook Comments)} Collected on socio-political issues of Bangladesh. Labelled into six fine-grained emotion classes: sadness, happiness, disgust, surprise, fear, and anger  & 6 & 5640 
& F1:62.39~\cite{parvin2021ensemble} \\
& \citet{tripto2018detecting} & \emph{(YouTube Comments)} Bangla, English, and Bangla (Romanized) YouTube comments from 2013 to 2018. Labelled into 3 and 5-class sentiment and 6 emotions & 3, 5, 6 & 15689 & F1:65.97 
 \\
& \citet{iqbal2022bemoc} & \emph{(BEmoC)} 7125 texts collected from Facebook, YouTube comments/posts, Bengali story books, etc. Labelled into six emotion categories: anger, fear, surprise, sadness, joy, and disgust by 5 annotators & 6 & 7000 & \textemdash \\ 
\midrule
{\multirow{8}{*}{\rotatebox[origin=c]{90}{\textbf{Hate Speech Detection}}}} & \citet{karim2021deephateexplainer} & \emph{(Facebook, YouTube, and Newspaper Comments)} Extended Bangla Hate Speech dataset~\cite{karim2020classification} for Personal, Geopolitical, Religious, etc. & 6 & 8087 
& F1:87 \\ 
& \citet{aurpa2022abusive}  & \emph{(Facebook Post)} Bangla texts containing cyberbullying~\cite{ahmed2021bangla} 
 from comments on Facebook posts of celebrities, athletes, and government officials. Labelled into five harassment categories: sexual, non-bullying, trolling, religious, and threats.  & 5  & 44001 
& F1:85.00 \\ \midrule
{\multirow{2}{*}{\rotatebox[origin=c]{90}{\textbf{DT}}}} & \citet{ghosh2023attention} & \emph{(Social Media)} 4784 Depressive and 10247 non-depression social texts collected and labeled as 0/1.  & 2  & 15031 
& A:94.32 \\
     \bottomrule
  \end{tabular}%
  }
\end{table*}

\section{Proposed Architectures}
As mentioned before, our aim in this survey is to discuss advances enabled by the Transformer architecture \cite{vaswani2017attention} and foundational variants originating from it. The Transformer model was introduced as a solution to bypass the recurrent language models of recent years such as ELMo \cite{peters-etal-2018-deep}. This is because, even though recurrent models were capable of capturing small-to-moderate length input dependencies, they were bottlenecked by their \textit{scalability issue} i.e. there was no way to parallelize computation. 

The Transformer then completely replaced recurrence with the \textit{attention} mechanism and feed-forward layers which enabled the model to be scaled up to parameters not seen before in deep learning literature. Essentially, a Transformer is a sequence-to-sequence (seq2seq) model consisting of an encoder, which ``encodes'' the input text to an internal \textit{contextualized} representation (using Multi-Headed Self Attention or MHA) and the decoder, conditioned on the encoder output, generates the target sequence (through masked MHA, as it is not allowed to look at tokens beyond its current timestep). 

The original Transformer was intended mainly for sequence transduction tasks (given an input, covert it to the relevant output). However, architectures started emerging based on two-phase training of either the encoder, decoder, or the entire seq2seq setup. In the first phase (\textit{pre-training}), the model is trained on a language modeling objective in a semi-supervised manner (the labels are obtained from the unlabelled corpus itself) followed by training (\textit{fine-tuning}) on a labeled downstream task, such as named entity recognition, SA, etc. by applying a task-specific layer (or \textit{head}) to the pre-trained checkpoint. Such two-phase training led to a slew of models which are popularly known today as \textit{Foundation Models} (FMs). 

\begin{figure*}
    \centering
    \includegraphics[width=0.8\textwidth]{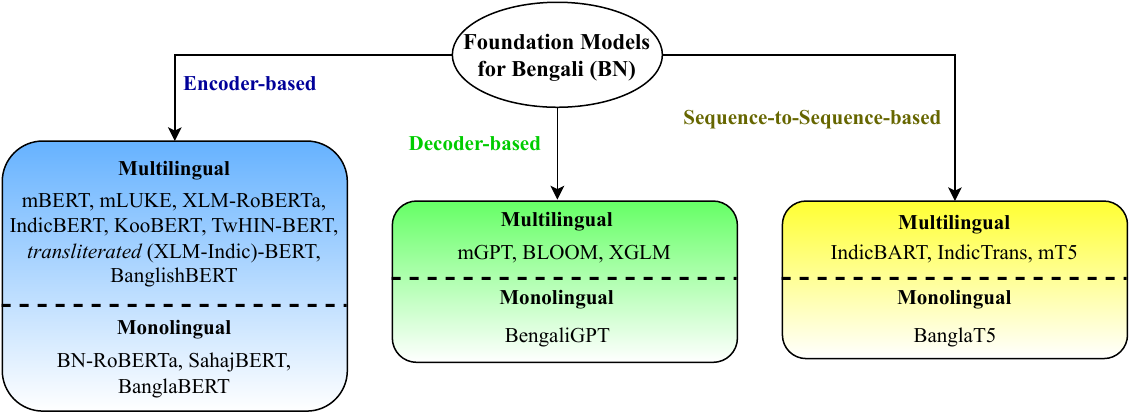}
    \caption{Models considered in this survey classified according to architecture type.}
    \label{fig:models}
\end{figure*}

In this work, we discuss only those architectures that are relevant to Bengali i.e. either have been pre-trained only on Bengali corpora or included as a part of the pre-training corpora (similar to the datasets, cf. section 2). With this in mind, we survey the models in figure \ref{fig:models}. Note that not all of these models were tested on SA or its related tasks. However, we organize them here to direct interested users to use them for SA since these models have seen Bengali during pre-training.

\subsection{Encoders}

In this section, we describe those FMs that use only the Transformer encoder. These models usually follow a \textbf{Masked Language Modelling} (MLM) objective as their main task along with auxiliary objectives such as entity prediction, etc. During MLM, a random number of tokens in the input sequence are \textit{corrupted} by means such as \textit{replacement (masking)} (replacing individual/spans of tokens by special ``mask''/random vocabulary tokens) or \textit{exclusion} (deleting a span) and the idea is to either predict the replaced/masked spans or the original sentence [Table 3 \cite{raffel2020exploring}]. We briefly discuss the relevant base architecture and models with Bengali as part of the training corpus, and/or additional training objectives.

\begin{enumerate}
    \item \textbf{BERT} \cite{devlin-etal-2019-bert} One of the first encoder-based models introducing the bidirectional attention mechanism, i.e. considering text from both sides of a token when computing its representation, along with the Next-Sentence Prediction (NSP) objective (to determine whether or not sentences A \& B follow each other) to allow the model to learn properties at the sentence level.
    \begin{itemize}
        \item[\ding{213}]\textcolor{blue}{M(multilingual) BERT}\footnote{HuggingFace model: \texttt{bert-base-multilingual-cased}} was simply pre-trained on 100 languages with the largest Wikipedia collections.
        
        \item[\ding{213}] \textcolor{blue}{KooBERT}\footnote{\url{https://huggingface.co/Koodsml/KooBERT}} was pre-trained on texts from the social media platform \textit{Koo India} \footnote{\url{https://www.kooapp.com/}}. The training corpus consists of 12 languages (including English) and we believe can be leveraged for useful \textit{social media text analysis} in Indian languages.

        \item[\ding{213}] \textcolor{blue}{TwHIN(Twitter heterogeneous information network) BERT} \cite{zhang2022twhin} was designed specifically for tasks involving social media data such as SA and hashtag prediction. TwHIN-BERT replaces the NSP objective with a task designed to determine if a pair of tweets are \textit{socially similar} or not, by computing a contrastive loss between each pair of tweets in a batch.
    \end{itemize}

    \item \textbf{ALBERT} \cite{DBLP:conf/iclr/LanCGGSS20} is an efficient variation of BERT as it is smaller in size (parameters) and faster to train. ALBERT achieves these improvements due to three architecture choices, a) sharing parameters across all Transformer blocks b) decomposing the embedding lookup table into two sub-matrices to change hidden dimension size without increasing the number of parameters drastically, and c) Considering Sentence-Order Prediction (SOP) over NSP. It is a type of NSP, but, the idea here is to determine whether a pair of sentences are in the correct ``order'' rather than whether sentence A ``follows'' B.

    \begin{itemize}
        \item[\ding{213}]\textcolor{blue}{IndicBERT} \citet{kakwani2020indicnlpsuite} pretrained an ALBERT-based model (IndicBERT) on their IndicCorp corpus which contains text in 12 languages (11 Indian + English) on topics of social interest such as online news sources and magazines. Curiously, they remove the SOP objective during pre-training relying only on MLM (reason not provided in their paper).

        \item[\ding{213}]\textcolor{blue}{SahajBERT} \citet{diskin2021distributed} proposed a novel training paradigm in which an ALBERT model was trained on Bengali corpora using a \textit{volunteer compute} approach i.e. pooling resources from several individuals to train a single model. This is a complete monolingual model consisting of the Bengali parts of the Wikipedia \footnote{\url{https://dumps.wikimedia.org}} and OSCAR \cite{ortiz-suarez-etal-2020-monolingual} corpus.

        \item[\ding{213}]\textcolor{blue}{Transliterated-BERT} Although not the official name, \cite{moosa2022does} propose ALBERT and RemBERT \cite{chung2020rethinking} models trained on \textit{transliterated} (writing a source language text in the target language script) versions of Indian languages. They conjectured that when several languages share similarities such as structure, common words, etc. it could be beneficial to transliterate them to a common script and train a FM on it. They chose a set of 20 languages (19 Indian + English) to transliterate to Latin script using the ISO 15919 standard and trained models on both the transliterated and multilingual corpus showing improvements with the former.
    \end{itemize}

    \item \textbf{RoBERTa} \cite{liu2019roberta} is essentially BERT but with better optimization methods such as dropping the NSP objective (which they show degrades performance), using byte-pair encoding instead of wordpiece and training with much larger data, batches and steps. This results in a model, larger (parameters \& vocabulary) \& more efficient than BERT. \textcolor{blue}{bnRoBERTa} \cite{jain2020indic} is a \underline{monolingual} RoBERTa model trained only on the Bengali portion of OSCAR while \textcolor{blue}{XLM-(R)oBERTa} \cite{conneau-etal-2020-unsupervised} is a \underline{multilingual} model trained on a dataset of 100 languages from CommonCrawl with other pre-training objectives such as Translation Language Modelling (concatenating sentences in two languages + MLM across all randomly masked tokens) for cross-lingual tasks such as XNLI, etc. \citet{yamada-etal-2020-luke} proposed \textbf{LUKE}, a RoBERTa model trained with an masked entity prediction task (since it treats ``words'' and ``entities'' as different) in addition to MLM. These additional objectives aid in entity-forward tasks such as NER, Cloze-style QA, etc. \textcolor{blue}{mLUKE} \cite{ri-etal-2022-mluke} then builds on LUKE by training a multilingual model (similar to XLM-R) on Wikipedia dumps of 24 languages.
    
    \item \textbf{ELECTRA} \cite{DBLP:conf/iclr/ClarkLLM20} is a generative adversarial network (GAN) like \textit{pre-training approach} in which two BERT models, termed generator and discriminator, are jointly trained. The generator is trained via standard MLM, producing outputs for the masked input sequence. After prediction, the masked tokens, in the original sequence, are replaced by sampling the generator output. The discriminator is then tasked to determine whether each token belongs to the original or replaced sequence. Such a setup has been shown to lead to a more efficient model in terms of training time and performance. \textcolor{blue}{BanglaBERT} and \textcolor{blue}{BanglishBERT} \cite{bhattacharjee-etal-2022-banglabert} are two ELECTRA models trained on monolingual and EN (same as English BERT's pre-training corpora)-BN corpora. Owing to the small size of the Bengali Wikipedia dump, the authors collected content from the top sources (news, blogs, ebooks, etc.) as cited by Amazon Alexa rankings. \textbf{To the best of our knowledge, BanglaBERT seems to be the ``best'' model} to date on a wide range of Bengali tasks (BLUB (Bangla Language Understanding Benchmark))
\end{enumerate}

\subsection{Decoders}

Unlike encoder models, Transformer decoders are more \textit{classical} in that they are trained using the extant paradigm of \textbf{CLM} (Causal Language Modelling) i.e. predicting the next word in a sequence being conditioned on the prior tokens, usually without auxiliary training objectives, unlike encoders. In decoders, the attention computation for a given token only has access to its preceding tokens (otherwise it would be cheating for the model to know what tokens come after it) unlike encoders which can use all of the tokens surrounding the masks. 

While there are several decoder-based architectures, the most popular in this category seems to be the \textbf{GPT} (Generative Pre-Training) family of models \cite{zong2022survey}. Each variant, GPT-1,2,3 etc. follows the training scheme for CLM but differs in ways such as using cleaner/diverse corpora and scaling up parameters resulting in better zero/few-shot capabilities. 

Although GPT-3 \cite{brown2020language} was shown to have reasonably good performance in languages apart from English \cite{armengol-estape-etal-2022-multilingual} the training data would suggest that for languages, like Bengali, occupying a very small fraction of the dataset, downstream performance would not be good. In an effort to combat this, \citet{lin-etal-2022-shot} proposed \textcolor{blue}{XGLM}, a multilingual generative model GPT-3 like model. Almost parallelly released was \textcolor{blue}{mGPT} \cite{shliazhko2022mgpt} a GPT-2 \cite{radford2019language} like architecture. While both are \underline{multilingual}, there exists subtle differences between the two. XGLM used GPT-3 curie and wanted to examine the effect of model scale for low-resource languages but trained on a smaller/focused set of the same (30 total). mGPT on the other hand, wanted to replicate the GPT-3 architecture using open-source tools such as the Megatron-LM framework \cite{shoeybi2019megatron}, perform comparison with XGLM and cover more languages (60). The only \underline{monolingual} GPT-based model seems to be \textcolor{blue}{Bengali-GPT} \cite{bangla-gpt2} a GPT-2 model trained on the Bengali part of the mC4 corpus \footnote{\url{https://huggingface.co/datasets/mc4}}. 

A surprise addition to this list of models is \textbf{BLOOM} \cite{scao2022bloom}, a BigScience  \footnote{\url{https://bigscience.huggingface.co/}} open-source multilingual model trained on scientific texts. Similar to mGPT, BLOOM uses Megatron's GPT-2 checkpoint to train on a diverse corpora of 46 different languages and 13 programming languages. Although their paper does not contain a breakdown of domains/topics considered per-language, we can see that with just 18 GB of Bengali corpora, they achieve impressive performance, at times comparable to high-resource languages like Spanish.

\subsection{Sequence-to-Sequence}

Finally, we discuss FMs consisting of encoders and decoders (seq2seq). In such setups, the encoder follows its MLM objective, formulates a representation for the input sequence, passes it over to the decoder which on being conditioned by it, performs CLM. However, during training, the decoder generates tokens in parallel irrespective of what the expected output is i.e. teacher forcing. 

For Indic languages, the first major contribution has to be \textcolor{blue}{IndicTrans} \cite{ramesh2022samanantar}, a Transformer trained on their \textit{Samanantar} dataset, a parallel corpus of 11 Indian languages to English. They mention that IndicTrans is a \textit{uniscript} model i.e. all non-English text is converted to the Devnagari script which allows for broader lexical coverage across all the Indian languages. 

\textbf{T5 (Text-To-Text Transfer Transformer)} \cite{raffel2020exploring} makes the case that all standard NLP tasks, such as classification, question answering, etc. can be cast as seq2seq problems. They train T5 using a combination of supervised (as above) and unsupervised (MLM) tasks. For the former, the input must be formatted to their requirements such as prefixing the input with the task handle, etc. while for masking in general, T5 uses \textit{span-masking} i.e. corrupting the input sequence by masking consecutive tokens, with the decoder tasked to predict the replaced spans. \textcolor{blue}{mT5} \cite{xue-etal-2021-mt5} is a multilingual-T5 model pre-trained on 101 languages from the mC4 corpus. It must be noted that since mT5 was trained using unsupervised CLM only, it will display random zero/few-shot performance, unless fine-tuned. \textcolor{blue}{BanglaT5} \cite{bhattacharjee2023banglanlg} is, to the best of our knowledge, the only monolingual seq2seq model for Bengali. Pre-trained on the \textit{Bangla2B+} corpus \cite{bhattacharjee-etal-2022-banglabert}, BanglaT5 achieves impressive performance in comparison to its other multilingual counterparts on their BanglaNLG benchmark.

Finally, we discuss \textbf{BART} \cite{lewis-etal-2020-bart} a seq2seq model trained \textbf{very similarly to T5} (span-masking). However, it is our understanding that apart from a few architecture choices such as absolute v/s relative position embeddings, the key difference between BART and T5 is that the latter was trained in a multi-task setting whereas the former was trained simply with MLM + CLM. The encoder's job is to reconstruct or \textit{denoise} the corrupted input sequence (by a variety of techniques such as masking, sentence reordering, etc.) while the decoder learns standard CLM using the original input sequence and encoder output. \textcolor{blue}{mBART-50} \cite{tang2021multilingual} is then a multilingual extension of BART, covering bilingual pairs of 49 languages to English and pre-trained on publicly available parallel corpora from WMT, IWSLT, etc. \textcolor{blue}{IndicBART} \cite{dabre-etal-2022-indicbart} is a further extension of mBART covering 11 Indic languages not seen by it earlier.

\section{Open Challenges} 

Seeing as SA in Bengali has come a long way, we highlight ongoing challenges in the field. As mentioned before, BN, like other low-resource languages, suffers from a \textbf{lack of resources}, datasets, and pre-trained models. Consider BERT for English. The corpora which was used, had a total of 3.3B tokens whereas the ``best'' model in Bengali, BanglaBERT, was trained with 2.1B tokens (\textbf{\textasciitilde1.5 times less}). Considering the linguistic complexity/diversity of BN, the value should have been 1.5 times more. This is a clear testament to the EN-BN performance gap. Efforts such as \textit{AI4Bharat}\footnote{\url{https://ai4bharat.org/}} are thus a brilliant initiative in this direction.

Considering the socio-geographic distribution of BN, we recognize that there are two camps of writers viz. those who write in pure BN script \& those who \textbf{code-mix} i.e write BN \& another language (typically EN) in the same sentence. This makes it difficult to develop models for BN as users comfortable in both languages typically fall back on EN through transliteration leaving pure-script users with underpowered models. However, with tools such as \textit{Google Keyboard} \footnote{\url{https://en.wikipedia.org/wiki/Gboard}} support for Bengali has become more engaging and easy \& with the appropriate permissions, we can curate a massive collection of user-generated pure-script text.

Mainly, there are two styles of BN, formal (\textit{sAdhu}) \& colloquial (\textit{calit}) \cite{pal2021search} and modern users typically favor the latter. This creates issues for models trained on more structured text, such as from Wikipedia, owing to most users relying on \textbf{relaxed grammar rules}. To alleviate this, models such as TwHIN-BERT should be developed for BN by training on a combination of both formal/colloquial and social-media texts. Additionally, as is a pervading issue in NLP nowadays, text mined from social media tends to be rife with biases. Thus, care needs to be taken to apply appropriate filtration before training our models.

\vspace{-0.3cm}
\subsection*{Future Directions}
While we motion for the need for larger and more diverse Bengali corpora, we recognize the difficulty of the task considering limitations in digitizing Bengali text such as scale, copyright issues and pipelines not being end-to-end (E2E) \cite{sankar2006digitizing}. However, we recognize an opportunity here from the ongoing challenges. If the majority of BN users prefer to transliterate their texts, why not take advantage of that? Developing \textbf{E2E transliteration-translation datasets} will be useful for training models to synthesize pure-script monolingual data.

Inspired by BLOOM, we encourage researchers to explore \textbf{complex domains} in BN such as Medical \cite{sazzed-2022-banglabiomed}, IT \cite{mumin2014sumono}, etc. Work along this direction can enable a wide range of applications in BN such as chat-agents capable of understanding both cultural nuances and medical jargon, aiding automated customer service etc.

Relying on text modality alone for detecting sentiment limits our models' ability to pickup on subtleties in communication such as prosody, body language, \cite{kundu2022survey}. Human beings express sentiment via several \textit{tells} such as facial expressions, voice modulation, etc. Looking beyond the extant text-based SA, a \textbf{multimodal} approach \cite{habimana2020sentiment, zhu2023multimodal} to the problem should be investigated. \textbf{Memes} provide the perfect hunting ground for such analysis. Not only are they a source of entertainment, the paradoxical nature of memes makes it difficult for our models to gauge the conveyed message, enabling us to determine their robustness. With the release of datasets as \textit{MemoSen} \cite{hossain-etal-2022-memosen}, research in this direction will be truly propelled.

\section{Conclusion}

In this paper, we have surveyed a wide array of Transformer-based models which can be used to address SA in Bengali. Going over the SOTA, it is clear that while positive strides have been made to push Bengali to high-resource territory, much work remains to be done. Highlighting the existing architectures and challenges in the field, we believe that our paper can be treated as a means to beckon more research in this area.

\section*{Limitations} 
In this paper, we focus our survey only on post-Transformer models, as other surveys \cite{shammi2023comprehensive} \cite{sen2022bangla} \cite{hira2022systematic} \cite{banik2019survey} \cite{alam2021review} on Bangla SA have covered pre-Transformer models extensively. We have also not independently verified claims made by the various papers meticulously since that goes beyond the scope of this paper. Our objective is to survey the state-of-the-art as it exists today in the literature assuming that the results claimed by the respective authors are reproducible.

\section*{Ethics Statement}
This work primarily summarizes the work done on the Bengali language and no additional work using any human input was obtained. We provide full credit to all original works and have tried to carefully quote all parts of this work that have been obtained verbatim from these works. We believe our paper will enable the world to better understand the state-of-the-art in Bangla NLP, specifically sentiment mining. The impact will be of convenience, i.e., a quicker understanding of the existing tools, and thus the same benefits and costs/abuses of those, apply here. However, the ethical risks arising from this paper is smaller since these techniques already exist. We merely make finding them easier.



\bibliographystyle{acl_natbib}
\bibliography{custom.bib}

\end{document}